\pdfoutput=1

\documentclass[11pt]{article}

\usepackage[]{acl}

\usepackage{times}
\usepackage{latexsym}

\usepackage{etoolbox}
\robustify\bfseries

\usepackage{siunitx}
\usepackage{booktabs}
\usepackage{multicol}
\usepackage{amssymb}
\usepackage{pifont}
\usepackage{xcolor} 
\usepackage{soul}
\usepackage{stfloats}

\usepackage[T1]{fontenc}

\usepackage[utf8]{inputenc}

\usepackage{microtype}
\usepackage{graphicx}
\usepackage{amsmath}
\usepackage[export]{adjustbox}
\usepackage{multirow}
\usepackage{caption}
\usepackage{subcaption}
\usepackage{placeins}

\newcommand{\cmark}{\ding{51}}%
\newcommand{\xmark}{\ding{55}}%
\newcounter{srCounter}
\newif\ifsrvar
\srvartrue
\ifsrvar
\newcommand{\seb}[1]{{\small \color{red} \refstepcounter{srCounter}\textsf{[SR]$_{\arabic{srCounter}}$:{#1}}}}
\else
\newcommand{\seb}[1]{}
\fi

\newcounter{fpCounter}
\newif\iffpvar
\fpvartrue
\iffpvar
\newcommand{\fabio}[1]{{\small \color{blue} \refstepcounter{fpCounter}\textsf{[FP]$_{\arabic{fpCounter}}$:{#1}}}}
\else
\newcommand{\fabio}[1]{}
\fi

\newcounter{mbCounter}
\newif\ifmbvar
\mbvartrue
\ifmbvar
\newcommand{\michele}[1]{{\small \color{purple} \refstepcounter{mbCounter}\textsf{[MB]$_{\arabic{mbCounter}}$:{#1}}}}
\else
\newcommand{\michele}[1]{}
\fi

\newcounter{syCounter}
\newif\ifsyvar
\syvartrue
\ifsyvar
\newcommand{\scott}[1]{{\small \color{violet} \refstepcounter{syCounter}\textsf{[SY]$_{\arabic{syCounter}}$:{#1}}}}
\else
\newcommand{\scott}[1]{}
\fi

\newcounter{plCounter}
\newif\ifplvar
\plvartrue
\ifplvar
\newcommand{\patrick}[1]{{\small \color{green} \refstepcounter{plCounter}\textsf{[PL]$_{\arabic{plCounter}}$:{#1}}}}
\else
\newcommand{\patrick}[1]{}
\fi

\newcounter{goCounter}
\newif\ifplvar
\plvartrue
\ifplvar
\newcommand{\giuseppe}[1]{{\small \color{brown} \refstepcounter{goCounter}\textsf{[GO]$_{\arabic{goCounter}}$:{#1}}}}
\else
\newcommand{\giuseppe}[1]{}
\fi

\newcommand{\eg}{\textit{e.g.}}
\newcommand{\ie}{\textit{i.e.}}
\newcommand{\system}{\textsc{SEAL}}

\title{Autoregressive Search Engines:\\ Generating Substrings as Document Identifiers}

\newcommand{\sapienza}{$^1$}
\newcommand{\sapienzafair}{$^{1,2}$}
\newcommand{\fair}{$^2$}
\newcommand{\uclfair}{$^{2,3}$}
\newcommand{\ucl}{$^3$}

\author{
Michele Bevilacqua\sapienzafair{} \
Giuseppe Ottaviano\fair{} \
Patrick Lewis\fair{} \\
{\bf Wen-tau Yih\fair{} \ Sebastian Riedel\uclfair{} \ Fabio Petroni\fair{} } \\ 
\sapienza{}Sapienza University of Rome \ \fair{}Meta AI \ \ucl{}University College London}

\begin{document}
\maketitle
\begin{abstract}
Knowledge-intensive language tasks require NLP systems to both provide the correct answer and retrieve supporting evidence for it in a given corpus. 
Autoregressive language models are emerging as the de-facto standard for generating answers, with newer and more powerful systems emerging at an astonishing pace.
In this paper we argue that all this (and future) progress can be directly applied to the retrieval problem with minimal intervention to the models' architecture. 
Previous work has explored ways to partition the search space into hierarchical structures and retrieve documents by autoregressively generating their unique identifier.
In this work we propose an alternative that doesn't force any structure in the search space: using all ngrams in a passage as its possible identifiers. This setup allows us to use an autoregressive model to generate and score distinctive ngrams, that are then mapped to full passages through an efficient data structure.
Empirically, we show this not only outperforms prior autoregressive approaches but also leads to an average improvement of at least 10 points over more established retrieval solutions for passage-level retrieval on the KILT benchmark, establishing new state-of-the-art downstream performance on some datasets, while using a considerably lighter memory footprint than competing systems. Code and pre-trained models at \url{https://github.com/facebookresearch/SEAL}.

\end{abstract}

\begin{figure*}[ht!]
    \centering
    \resizebox{\textwidth}{!}{%
    \includegraphics[
    trim={0cm 0cm 0cm 0cm},    
    clip=true
    ]{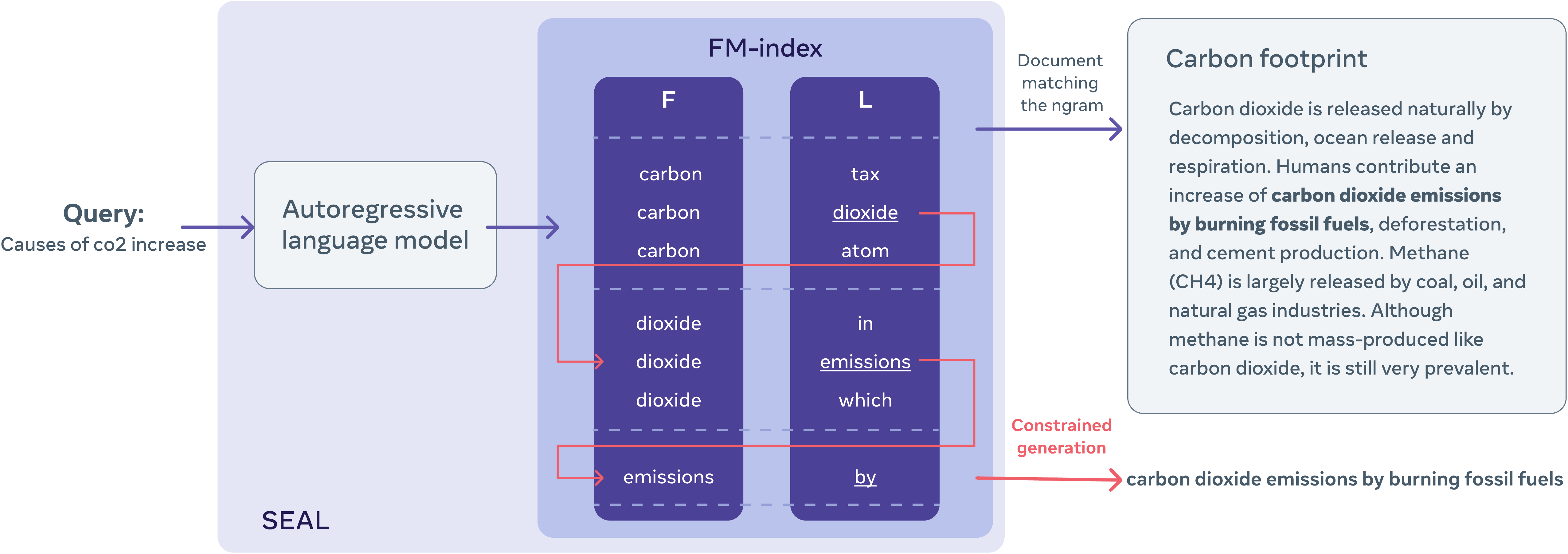}}
    \caption{
    High-level SEAL architecture, composed of an autoregressive LM paired with an FM-Index, for which we show the first (F) and last (L) columns of the underlying matrix (more details in Sec \ref{sec:fm-index}). The FM-index constraints the autoregressive generation (\eg, after \textit{carbon} the model is contrained to generate either \textit{tax}, \textit{dioxide} or \textit{atom} in the example) and provides the documents matching (\ie, containing) the generated ngram (at each decoding step).
    }
    \label{fig:main}
\end{figure*}

\section{Introduction}

Surfacing knowledge from large corpora is a crucial step when dealing with knowledge intensive language tasks~\cite{levy2017zero,dinan2018wizard,elsahar2019t,petroni-etal-2021-kilt}, such as open-domain question answering~\cite{voorhees1999trec,joshi2017triviaqa,yang2018hotpotqa,kwiatkowski-etal-2019-natural} and fact checking~\cite{Thorne18Fever}. 
A  popular paradigm to approach such tasks is to combine a search engine with a machine reader component. The former retrieves relevant context, usually in the form of short passages, which the latter then examines to produce answers~\cite{DBLP:journals/corr/ChenFWB17,lewis2020retrievalaugmented, izacard-grave-2021-leveraging}. 

In recent years we have witnessed a surge of research and development in autoregressive language models~\cite{radford2019language, Lewis2019BARTDS, 2019t5, brown2020language, rae2021scaling, artetxe2021efficient, smith2022using}, with ever increasing size and natural language understanding (NLU) capabilities. Such models are currently the de-facto implementation of the machine reader component in retrieval-reader architectures, and have contributed to rapid progress on a wide range of benchmarks \cite{joshi2017triviaqa,kwiatkowski-etal-2019-natural, petroni-etal-2021-kilt}. 
However, these tremendous advances in aggressive modelling has yet to bring similar transformational changes in  how retrieval is approached.

Transferring the NLU capabilities of modern autoregressive models to retrieval  is non-trivial. 
Some works have demonstrated that knowledge stored in the parameters of these models can be retrieved to some extend by directly generating evidence given a query \cite{petroni2019language, petroni2020context, roberts2020much}.
However, such approaches have been shown to be unreliable because of their tendency to hallucinate non-factual content \cite{massarelli2019decoding, Metzler:Rethinking21, Ji2022SurveyOH}.
To alleviate this issue, previous work proposes to only use generation for query expansion in traditional search engines \citep{mao-etal-2021-generation}, but these solutions don't exploit the full potential of autoregressive architecture, such as word order sensitivity and conditional probability modeling, and still lag behind vector-based approaches~\cite{karpukhin-etal-2020-dense}.

Recently, another line of work has investigated using autoregressive language models to generate identifier strings for documents, as an intermediate target for retrieval, such as Wikipedia page titles \citep{decao-etal-2021-autoregressive}, or root-to-leaf paths in a hierarchical cluster tree \citep{tay-etal-2022-transformer}. 
Employing identifiers, rather than generating evidence directly, induces some structure in the search space, (\ie, index documents by their title or their cluster tree) which can be easier to memorize, learn, and retrieve from, than full unstructured passages.
Moreover, it is relatively easy to constrain beam search decoding with a prefix tree ``index'' so that only valid identifiers are generated. As a downside, if appropriate metadata (e.g. titles) are not available, one needs to create the identifiers, hence the structure (e.g. with hierarchical clustering), which has not been thoroughly evaluated on a large-scale benchmark.

In this work, we propose a solution that does not force any structure in the search space, but rather uses all the ngrams occurring in a document as its identifiers.
Concretely, we introduce \textbf{S}earch \textbf{E}ngines with \textbf{A}utoregressive \textbf{L}Ms (\system{}), a retrieval solution that combines an autoregressive model, \ie, BART \citep{Lewis2019BARTDS}, with a compressed full-text substring index, \ie, the FM-Index \citep{ferragina-manzini-2000-opportunistic} --- see Figure \ref{fig:main} for an high-level overview. 
This configuration comes with a twofold benefit: i) we can constrain BART's generations with the FM-Index, hence preventing the generation of invalid identifiers (\ie, ngrams not occurring in any document); ii) the FM-Index provides information on all documents in the corpus containing a specific ngram (for every decoding step), thus allowing to retrieve them.
This setup allows \system{} to generate \emph{any span} from \emph{any position} in the corpus, without needing to explicitly encode all substrings in a document.
Moreover, we design a novel scoring function to intersect the results of multiple ngrams combining LM probabilities with FM-index frequencies (\ie, number of occurrences of the ngram in the whole corpus).%

Our experimental evaluation shows that \system{} matches or outperforms recent retrieval solutions (including autoregressive ones) on Natural Questions \cite{kwiatkowski-etal-2019-natural}, while requiring substantially less memory ($\sim$2 to 7 times smaller in footprint). Moreover, \system{}'s intersection formulation improves the state-of-the-art on passage-level retrieval by more than 10 points on the KILT benchmark \cite{petroni-etal-2021-kilt}, contributing in establishing new state-of-the-art downstream results on multiple datasets when paired with existing reader technologies.

\section{Related Work}

\paragraph{Retrieval with Identifiers} One way to approach retrieval with autoregressive models makes use of identifiers, \ie, string pointers to documents that are in some way easier to generate than the full document itself. In tasks where such data is available or relevant, such as Wikipedia-based entity linking (a form of page-level retrieval), titles have been shown to work well as identifiers~\citep{decao-etal-2021-autoregressive,de-cao-etal-2021-highly,de2022multilingual}. However, even on Wikipedia-based benchmarks, titles on their own are not well-suited for retrieval at passage-level, given they can only identify an article (that might contain several passages). In a different direction, \citet{tay-etal-2021-measuring} have used hierarchical clustering on contextualized embeddings to create identifiers for arbitrary spans of text. In contrast, in our work the identifiers are corpus string matches, which do not necessarily occur in just one document.

\paragraph{Term Weighting}
Virtually all modern approaches to string-matching-based sparse retrieval make use of a bag-of-words assumption, indexing documents with an \emph{inverted index}, a data structure mapping terms to documents or, more generally, locations in a corpus~\citep{robertson-zaragoza-2009-probabilistic}.
Retrieval performance in this setting depends heavily on term-weighting schemes, with many recent works proposing sophisticated, contextualized weights for both queries, and documents\citep{dai-callan-2019-context,gao-etal-2021-coil,lin-ma-2021-unicoil,mallia-etal-2021-deepimpact,dai-callan-2020-context,bai-etal-2020-sparterm,zhao-etal-2021-sparta,formal-etal-2021-splade,formal-etal-2021-spladev2}.
Many of these methods are also able to weigh terms that are not present in the query, addressing so-called vocabulary mismatch. In contrast, \system{} generates (and assigns scores to) ngrams of arbitrary size, using the index for both generation and retrieval. Nevertheless, this line of work is partly orthogonal to our own, as many of the proposed techniques could be used to rescore higher-order ngrams.

\paragraph{Query/Document Expansion}
A line of research which often involves autoregressive language models is that of document and query expansion. For example, one can augment stored documents by generating possible queries that might be answered by them~\citep{nogueira-etal-2019-doc2query,nogueira-lin-2019-docTTTTTquery}. 
In the opposite direction, works like GAR~\citep{mao-etal-2021-generation} augment the query by predicting helpful additional terms, such as an answer, sentence containing the answer, or the title of a document where the answer may be found.
We note that while query expansion bears a superficial resemblance with \system{}, the approaches are  conceptually distinct. 
While query expansion methods rely on a stand-alone black-box retriever, in our work the boundary between generation and retrieval is blurred, since our identifiers are grounded passage spans. 

\paragraph{Query Likelihood Models}
Another connected strand of research is that of query likelihood models, which, in their latest incarnations, use autoregressive models to (re)rank passages according to the probability $P(q|p)$ of a query $q$ given the passage $p$~\citep{nogueira-dos-santos-etal-2020-beyond,zhuang-zuccon-2021-tilde,lesota-etal-2021-modern}. In our case, the autoregressive architecture models the likelihood of an ngram given the query, \ie, $P(n|q)$. 

\paragraph{``Learning to Google''}
Recently, language models have been shown to be able to directly generate search queries for modern web search engines either with finetuning on demonstrations~\citep{Komeili2021InternetAugmentedDG,Shuster2022LanguageMT} and human preferences~\citep{nakano-2021-webgpt} or via prompting~\citep{lazaridou-etal-2022-internet}. In our case, there is no black-box retrieval system that is queried. Rather, the white-box index determines both the generated ngrams and the search process.

\section{Background}

In retrieval, the automatic system is required to return an ordered list of documents $d_1, d_2, \dots, d_n$ from a retrieval corpus $\mathcal{R}$, given a query $q$. Both queries and documents are texts, \ie, lists of tokens $\langle t_1, t_2, \dots, t_N\rangle$, where each token $t$ is drawn from a vocabulary $V$. A span of tokens in a text is called an ngram; ngrams of size $1$ are known as unigrams. We denote with $F(n, \mathcal{R})$ the frequency of an ngram $n$ in $\mathcal{R}$, \ie, the total number of times it appears in the whole retrieval corpus.

\subsection{The FM-Index}
\label{sec:fm-index}
Our method requires a data structure that can support the efficient identification of occurring substrings to guarantee that all decoded sequences are located somewhere in the retrieval corpus. Moreover, to perform retrieval, we require the ability to identify which documents the generated ngrams appear in. 
Neither inverted indices, (which have no efficient way to to search for phrases of arbitrary length),
nor prefix trees, (which would force us to explicitly encode all $k$ suffixes in a document), are viable options. 
The core data structure that satisfies our requirements is the 
FM-index~\citep{ferragina-manzini-2000-opportunistic}, \ie, a compressed suffix array that, as a self-index, requires no additional storage for the original text. FM-index space requirements are linear in the size of the corpus, and, with small vocabularies such as those used by modern subword-based language models, is thus usually \emph{significantly smaller} than the uncompressed corpus.
The FM-index can be used to count the frequency of any sequence of tokens $n$ in $O(|n| \text{log}|V|)$, \ie, independently from the size of the corpus itself. For constrained decoding, the list of possible token successors can be obtained in $O(|V| \text{log}|V|)$.
Internally, the FM-index relies on the Burrows-Wheeler Transform~\citep{Burrows94ablock-sorting}, or \emph{BWT}, an invertible transformation that permutes a string to make it easier to compress, defined as follows: all the rotations of the string are sorted lexicographically and laid out in a matrix; the last column of the matrix is the strings's BWT.\footnote{Since our corpus contains multiple documents, we concatenate them with a separator token.} For example, given the string $CABAC$, the corresponding matrix would be:
\begin{equation}
\nonumber
\footnotesize
\begin{matrix}
\textbf{F}  &   &   &   &   & \textbf{L} \\
\$^6 & C & A & B & A & C^5 \\
A^2 & B & A & C & \$ & C^1 \\
A^4 & C & \$ & C & A & B^3 \\
B^3 & A & C & \$ & C & A^2 \\
C^5 & \$ & C & A & B & A^4 \\
C^1 & A & B & A & C & \$^6 \\
\end{matrix}
\end{equation}

\noindent where $\$$ is a special end-of-string token. 
The first (\textbf{F}) and last (\textbf{L}) columns are the only ones that will be explicitly stored in the FM-index; \textbf{F} is just an array of runs (\ie, sequences of repeated tokens), due to the rotations being sorted, so it can be represented with one count for each alphabet symbol; \textbf{L}, the string's BWT, will be stored in a data structure known as the Wavelet Tree~\citep{grossi-etal-2003-highorder}, which allows efficient rank-select-access queries, while exploiting the compressibility induced by the transformation.
FM-indices have the useful property that for each symbol, the relative rank stays the same: that is, the $i$th occurrence of a symbol $\sigma$ in \textbf{F} points to the same location in the corpus of the $i$th occurrence of $\sigma$ in \textbf{L}. Thanks to this property, we can locate any string $\langle \sigma_1 , \sigma_2 , \dots , \sigma_n \rangle$ in the index by starting from $\sigma_n$ and going backwards. First, we select the contiguous range of rows corresponding to the symbol $\sigma_n$ in \textbf{F}, then we check the ranks of the first and last occurrences of $\sigma_{n-1}$ in the same range of rows in $L$. We use the ranks to select a new, smaller or equal range of rows looking up the symbol $\sigma_{n-1}$ in $F$. The procedure can be applied iteratively to find ngrams of arbitrary size.

\section{Method}
\label{sec:method}
In our retrieval methodology, \system{}, we generate multiple ngrams, conditioning on a query. The ngrams are then used to find the documents they appear in within the corpus, which are then returned to the user. In Figure \ref{fig:main} we show this process at a high-level. We use our indexing structure, \ie, the FM-index to constrain decoding so that each ngram occurs at least once in the retrieval corpus. Jointly, we use the FM-index to efficiently find matching documents. Documents are ranked using the scores of the generated ngrams. 

\paragraph{Autoregressive Retrieval}
\label{ssec:generating}

We generate ngrams identifiers with constrained beam search, using the FM-index to identify the set of possible next tokens in at most $O(|V| \text{log}|V|)$: tokens corresponding to unattested continuations are blocked by masking the logit to $-\infty$.
As a result, after a single decoding pass, we get a set of ngrams ($K$), along with their autoregressively-computed probabilities according to the model. It is also trivial to find the positions in the corpus where the decoded ngrams appear, as constrained decoding already requires selecting the relevant range of rows in the FM-index.
Note that autoregressive scoring entails monotonically decreasing scores---any string will be assigned a lower probability than any of its prefixes. To address this issue, we use fixed-length ngrams. Each document is assigned the score ($P(n|q)$) of its most probable decoded occurring ngram. 
We refer to this as the \textbf{LM} scoring. 

\paragraph{Factoring in FM-index frequencies}
To counterbalance the monotonic probability decrease, we integrate in scoring unconditional ngram probabilities, computed as normalized index frequencies:
\begin{equation}
    P(n) = \frac{F(n, \mathcal{R})}{\sum_{d \in \mathcal{R}} |d|}
\end{equation}
This also enables us to promote \textit{distinctive} ngrams, \ie, those that have high probability according to the model and low probability according to the FM-index. We take inspiration from the theory behind TF-IDF and BM25~\citep{robertson-zaragoza-2009-probabilistic} and use the following scoring function:

\begin{equation}
    w(n,q) = \text{max}( 0, \log \frac{P(n|q)(1 - P(n))}{P(n)(1 - P(n|q))})
\label{eq:ngram-score}
\end{equation}

\noindent
This formulation addresses the problem of length, as the unconditional probability of an ngram will also be equal or lower than that of any of its prefixes.
To make better use of the computational resources, we slightly modify the beam search implementation to keep track of all the partially decoded sequences that have been considered. Thanks to this, we score a larger number of ngrams than the size of the beam.
We refer to this formulation as the \textbf{LM+FM} scoring.

\paragraph{An Intersective Scoring for Multiple Ngrams}
One problem with the previous scoring formulations is that it is impossible to break ties among documents whose highest scoring ngram is the same, as they receive exactly the same score. Moreover, it might be difficult to capture all relevant information within a document by considering only a single ngram, for instance when salient ngrams are non-contiguous (\eg, separated by unrelated text).  
To address these issues we propose a novel scoring formulation that aggregates the contribution of multiple ngrams contained in the same document.
To avoid repeated scoring of overlapping ngrams, for each document $d \in \mathcal{R}$ we only consider a subset of the generated ngrams $K^{(d)} \subset K$. An ngram $n$ belongs to $K^{(d)}$ 
if there is at least one occurrence of $n$ in $d$ that does \textit{not} overlap with an occurrence of another ngram $n'$ such that a) $n' \in K^{(d)}$ b) $w(n', q) > w(n, q)$.
The document-level score, then, is the weighted sum of all ngrams in $K^{(d)}$:
\begin{equation}
W(d,q) = \sum_{n \in K^{(d)}} w(n, q)^\alpha \cdot \text{cover}(n, K^{(d)})
\label{eq:doc-score}
\end{equation}

\noindent
where $\alpha$ is a hyperparameter and the weight $\text{cover}(n, K)$ (controlled by the second hyperparameter $\beta$) is a function of how many ngram tokens are not included in the coverage set $C(n, K) \subset V$, \ie, the union of all tokens in ngrams with a higher score. We define this coverage weight as follows: 

\begin{equation}
    \text{cover}(n, K) = 1 - \beta + \beta \cdot \frac{|\text{set}(n) \setminus C(n, K) |}{|\text{set}(n)|}
\label{eq:cov-score}
\end{equation}
\noindent
The purpose of the coverage weight is to avoid the overscoring of very repetitive documents, where many similar ngrams are matched. Note that by saving the probability distribution at the first decoding step we can compute scores for all unigrams with no additional forward pass.
We refer to this last approach, which can be thought of as a higher-order generalization of the bag-of-words assumption, as the \textbf{LM+FM intersective} scoring.

\section{Experimental Setting}
Our experimental setting evaluates \system{} on English knowledge-intensive NLP tasks. Each considered dataset is a collection of queries, each of which can be answered by looking for piece(s) of evidence in the corpus. We consider both an in vivo evaluation, in which we assess the model by looking at how well the document ranking matches with the ground truth, and, in addition, we perform a downstream evaluation, in which we feed the retrieved documents to a trained reader, that uses the documents to generate the answer.

\subsection{Data}
\label{sec:exp-data}

\paragraph{Natural Questions}
Natural Questions (NQ) is dataset containing query-document pairs, where the query is a question (\eg, ``who wrote photograph by ringo starr''), and the document is a Wikipedia page, in which a span is marked as an answer~\citep{kwiatkowski-etal-2019-natural}. We experiment on both the customary retrieval setup used by, among others,~\citet{karpukhin-etal-2020-dense} and~\citet{mao-etal-2021-generation}, and the substantially different setup used by~\citet{tay-etal-2022-transformer}. We refer to these two settings as, respectively, \textbf{NQ} and \textbf{NQ320$k$}. In NQ, retrieval is performed on an entire Wikipedia dump, chunked in around $21$M passages of 100 tokens. Performance is measured as accuracy@$k$, \ie, the fraction of instances for which at least one of the top-$k$ retrieved passages contains the answer. NQ320$k$ is a much more restricted setting, in which the retrieval set is limited to the union of all ground truth document in the training, dev or test set. Different revisions of the same Wikipedia page count as different documents. Note that the exact splits used by~\citet{tay-etal-2022-transformer}, the retrieval corpus and the preprocessing code have not been yet released at the time of writing. Therefore, we have tried to replicate the setting as closely as possible, 
 but the exact numbers are not precisely comparable with those reported in the original paper. In NQ320$k$, performance is measured as hits@$k$, i.e, the fraction of instances for which at least one of the top-$k$ retrieved passages is in the ground truth.

\paragraph{KILT} is a comprehensive benchmark collecting different datasets including question answering, fact checking, dialogue, slot filling, and entity linking~\citep{petroni-etal-2021-kilt}. All these tasks are solvable by retrieving information from a unified corpus --- a Wikipedia dump. In KILT, the evidence is usually the paragraph that contains the answer. Following~\citet{maillard-etal-2021-multi}, we have re-chunked KILT's retrieval corpus, which is originally paragraph-based, in around $36$M passages of 100 tokens. We do not use the entity linking and  ELI5 KILT tasks, where a ground truth passage is not provided in the training set. KILT's retrieval performance is measured with R-precision, a precision-oriented measure that considers only gold documents as correct answers, not just any document containing the answer. R-precision can be computed at either passage level or at page level.

\subsection{\system{} configuration}

\paragraph{Training}
We finetune BART large~\citep{Lewis2019BARTDS} to generate ngrams of length $k=10$ from the ground truth document. Since there are $|d|-k$ ngrams in a document $d$, we sample (with replacement) $10$ ngrams from it, biasing the distribution in favor of ngrams with a high character overlap with the query. We also add the title of the document to the set of training ngrams. To expose the model to more possible pieces of evidence, we also add  different ``unsupervised'' examples for each document in the retrieval corpus to the training set. In each of these examples the model takes as input a uniformly sampled span from the document, and predicts either another sampled span, or the title of the page. 
We append special tokens to the input to signal to the model a) whether the pair comes from the supervised or unsupervised training pairs (in the same spirit as the co-training task prompts used by~\citet{tay-etal-2022-transformer}) b) whether a title or span is expected as output. On KILT we train \system{} on all datasets at once.

\paragraph{Training Hyperparameters}
We finetune the model using \texttt{fairseq}. We use Adam~\citep{DBLP:journals/corr/KingmaB14} with a learning rate of $3 \cdot 10^{-5}$, warming up for $500$ updates, then using polynomial decay for at $800k$ updates, evaluating every $15k$ steps. We stop the training run if the loss on the development set stops improving for $5$ evaluation passes. We use label smoothing ($0.1$), weight decay ($0.01$), and gradient norm clipping ($0.1$). We train in batches of $4096$ tokens on $8$ GPUs. 

\paragraph{Index}
We use the C++ FM-index implementation in \texttt{sdsl-lite}.
While the FM-index construction (which requires a sort of all rotations) takes around 6 hours in our single-threaded implementation, parallel algorithms are available~\citep{LABEIT20172}. Each document is encoded as the subword tokenization of the concatenation of the title and the passage, separated by a special token. We report in Table \ref{tab:nq-size} the index statistics for Natural Questions. As can be seen, \system{}'s FM-index is more than 7 times lighter compared to DPR's full document embeddings for exact inner product search, and needs neither a GPU for search on top of that, nor separate storage for the text itself. While vector compression methods can reduce dense retrievers' index size, this still comes at the expense of performance~\citep{yamada-etal-2021-efficient,Lewis2021BoostedDR}. In addition, our the size of  our index is less than 50\% of that of the well-optimized Lucene BM25 index used by \texttt{pyserini}, but also roughly 65\% of the uncompressed plain text itself.
\begin{table}[t]
\centering
\resizebox{0.85\columnwidth}{!}{%
\begin{tabular}{lcccc}
\toprule
\multirow{2}{*}{\textbf{System}} & 
\multirow{1}{*}{\textbf{Model}} & 
\multicolumn{3}{c}{\textbf{Index}} \\
& {\textbf{Params}} & {\textbf{Size}} & {\textbf{Params}} & {\textbf{GPU?}} \\ \midrule
\textit{plain text} & {-} & 13.4GB & {-} & {-}\\
\midrule
DPR      & 220M   & 64.6 GB    & 16.1B        & \cmark \\
BM25     & -      & 18.8 GB    & {-}          & \xmark \\
GAR      & 406M   & 18.8 GB    & {-}          & \xmark \\
DSI-BART & 406M   & {-}        & {-}          & {-}    \\
SEAL     & 406M   & 8.8GB       & {-}          & \xmark \\
\bottomrule
\end{tabular}
}
\caption{Language model and index size on Natural Questions (around 21M passages). \system{}'s index is \textasciitilde1.5 times smaller than uncompressed plain text.}
\label{tab:nq-size}
\end{table}

\paragraph{Inference} We decode for $10$ timesteps with a beam size of $15$, and set the hyperparameters $\alpha$, and $\beta$ to, respectively, $2.0$ and $0.8$. The hyperparameters have been tuned on the Natural Questions development set (§\ref{sec:exp-data}). In the constrained decoding stage, we force part of the generated ngrams to match document titles. 

\begin{table}[t]
\footnotesize
\centering

\begin{tabular}{l|SS}
\toprule

\multirow{2}{*}{\textbf{System}} & \multicolumn{2}{c}{\textbf{hits@\textit{k}}} \\
 & {\textbf{1}} & {\textbf{10}} \\

\midrule
BM25 (\texttt{gensim}) & 15.3 & 44.5 \\
BM25 & 22.7 & 59.0 \\
DSI-BART  & 25.0 & 63.6 \\
GENRE & \bfseries 26.3 & 71.2 \\
\midrule
\system{} (LM, $|n|=3$) & 21.3 & 66.5 \\
\system{} (LM, $|n|=4$) & 22.2 & 68.2 \\
\system{} (LM, $|n|=5$) & 22.6 & 68.7 \\
\system{} (LM+FM)           & 25.3 & 72.0 \\
\system{} (LM+FM, intersect.)      & \bfseries 26.3  & \bfseries 74.5 \\
\bottomrule
\end{tabular}
\caption{Results on NQ320$k$. Reporting hits@1 and hits@10. Best in bold.}
\label{tab:dsi-main}
\end{table}
\begin{table*}[ht]
\footnotesize
\centering

\begin{tabular}{lSSS|SSSS|S}
\toprule
\multirow{2}{*}{\textbf{System}} & \multicolumn{3}{c|}{\textbf{accuracy@\textit{k}}} & \multicolumn{4}{c|}{\textbf{Overlap? (A@100)}} & {\multirow{2}{*}{\textbf{EM}}} \\
 & {\textbf{5}} & {\textbf{20}} & {\textbf{100}} & {ans. \cmark } & {\xmark} & {ques. \cmark} & {\xmark} & \\
\midrule
BM25 & 43.6 & 62.9 & 78.1 & 82.9 & 70.1 & 80.9 & 76.6 &  40.42 \\
DPR \citep{karpukhin-etal-2020-dense} & \bfseries 68.3  & \bfseries 80.1 & 86.1 & 91.4 & 76.8 & 93.2 & 83.2 & 47.2 \\ 
GAR \citep{mao-etal-2021-generation} & 59.3 & 73.9 & 85.0 & \bfseries 91.6 & 74.4 & \bfseries 94.1 & 80.4 & 46.15 \\
DSI-BART & 28.3 & 47.3 & 65.5 & 77.8 & 44.2 & 84.9 & 57.7 & 31.4 \\
\citet{izacard-grave-2021-leveraging} & {-} & {-} & {-} & {-} & {-} & {-} & {-} & \bfseries 48.2 \\
\midrule
\system{} (LM, $|n|=5$) & 40.5 & 60.2 & 73.1 & 82.2 & 57.1 & 85.2 & 64.9 & 36.0 \\
\system{} (LM+FM) & 43.9 & 65.8 & 81.1 & 86.9 & 70.9 & 89.5 & 78.1 & 42.9 \\
\system{} (LM+FM, intersective) & 61.3 & 76.2 & \bfseries 86.3 & 91.2 & \bfseries 77.7 & 93.2 & \bfseries 84.1 & 48.0 \\
\bottomrule
\end{tabular}
\caption{Retrieval results on the NQ test set. Column blocks (left to right): retrieval results (accuracy@5/20/100); retrieval results on the test splits of \citet{lewis-etal-2021-question}, partitioned according to whether the query/answer is a paraphrase of one in the training set; downstream performances (exact match). Except for \citet{izacard-grave-2021-leveraging}, all downstream results are computed with the same FiD reader trained on DPR predictions. Best in bold.}
\label{tab:vanilla-nq}
\end{table*}
\begin{table*}[ht]
\footnotesize
\setlength{\tabcolsep}{5pt}
\centering
\begin{tabular}{lSSSSSSS|S}
\toprule
\textbf{Model}        & {\textbf{FEV}}  & {\textbf{T-REx}} & {\textbf{zsRE}} & {\textbf{NQ}}   & {\textbf{HoPo}} & {\textbf{TQA}}  & {\textbf{WoW}}  & {\textbf{AVG}}  \\ \midrule
BM25                & 40.1 & 51.6  & 53.0 & 14.2 & 38.4 & 16.2 & 18.4 & 33.1 \\
DPR \cite{maillard-etal-2021-multi}          & 43.9 & 58.5  & \bfseries 78.8 & 28.1 & 43.5 & 23.8 & 20.7 & 42.5 \\
MT-DPR \citep{maillard-etal-2021-multi}       & 52.1 & 53.5  & 41.7 & 28.8 & 38.4 & 34.2 & 24.1 & 39.0 \\
MT-DPR \citep{oguz-etal-2021-domain}      & 52.1 & \bfseries 61.4  & 54.1 & 40.1 & 41.0 & 34.2 & 24.6 & 43.9 \\ \midrule
MT-DPR$\dag$ \citep{oguz-etal-2021-domain}                      & 61.4 & 68.4  & 73.3 & 44.1 & 44.6 & 38.9 & 26.5 & 51.0 \\
MT-DPR$\dag$ (large) \citep{oguz-etal-2021-domain}     & 62.8 & 66.6  & 66.9 & 42.6 & 42.1 & 37.9 & 23.4 & 48.9 \\ \midrule

\system{} (LM+FM)   & 31.5 & 42.0 & 34.0 & 21.7 & 24.7 & 21.4 & 17.6 & 27.6 \\ 
\system{} (LM+FM, intersective)   & \bfseries 67.8 & 58.9 & \bfseries 78.8 & \bfseries 43.6 & \bfseries 54.3 & \bfseries 41.8 & \bfseries 36.0 & \bfseries 54.5 \\
\bottomrule

\end{tabular}
\caption{Retrieval results on individual KILT dev set(s), with the average in the rightmost column. Reporting passage-level R-precision (higher is better). We mark model that are also trained on additional synthetic data \citep{lewis-etal-2021-paq} with $\dag$. All \system{} models are multitask. Best among models trained only on KILT queries in bold.}
\label{tab:kilt-dev-rp}
\end{table*}

\subsection{Retriever Baselines}

We compare \system{} against well-established systems in the literature on each benchmark. On NQ and NQ320$k$ we also compare against our BART-based replication of DSI~\citep[\textbf{DSI-BART}]{tay-etal-2022-transformer}. On NQ320k, a page-level benchmark, we include our own replication of GENRE~\citep{decao-etal-2021-autoregressive}. Unless otherwise specified, we use \texttt{pyserini} to compute the BM25 baseline. For other systems, we either take figures from the literature, or use publicly released model predictions.

\paragraph{DSI-BART} On NQ320$k$, \texttt{bert-base-cased} is used to compute the embeddings for the clustering. On regular NQ, we use the public precomputed DPR embeddings. To compare fairly against \system{}, we fine-tune the same encoder-decoder backbone, \ie, BART large.

\subsection{Reader}
For downstream results, we use the Fusion-in-Decoder abstractive reader~\citep{izacard-grave-2021-leveraging}, which takes in the query along with 100 contexts and produces a task-specific answer. We train FiD on training set predictions.

\section{Results}
\paragraph{NQ320$k$}

We report results on NQ320$k$ in Table \ref{tab:dsi-main}. 
\system{} outperforms BM25 and DSI-BART in hits@10 in all its formulations. When taking into account ngram frequencies (\ie, LM+FM), \system{} achieves even higher results than GENRE, despite the fact that this benchmark only requires page-level retrieval capabilities (that is the focus of GENRE).  Finally, our intersective formulation achieves the highest results, both in hits@1 and @10, indicating that multiple ngrams identifiers might capture complementary information, which can be aggregated for stronger performances.

\begin{table*}[!ht]
\footnotesize
    \centering
    \begin{tabular}{lSSSSSSS}
    \toprule
    \multirow{2}{*}{\textbf{System}}         & {\textbf{FEV}}  & {\textbf{T-REx}} & {\textbf{zsRE}} & {\textbf{NQ}}   & {\textbf{HoPo}} & {\textbf{TQA}}  & {\textbf{WoW}}  \\ 
                   & {\textsc{acc}}  & {\textsc{acc}}   & {\textsc{acc}}  & {\textsc{em}}   & {\textsc{em}}   & {\textsc{em}}   & {\textsc{f1}}   \\ \midrule
    KGI \citep{glass-etal-2021-robust}$^\dag$          & 85.6 & \bfseries 84.4  & 72.6 & 45.2 & {-} & 61.0 & 18.6 \\
    Hindsight \citep{paranjape-etal-2021-hindsight}      & {-} & {-} & {-} & {-} & {-} & {-} & \bfseries 19.2 \\ 
    DPR+BART \citep{petroni-etal-2021-kilt}        & 86.7 & 59.2  & 30.4 & 41.3 & 25.2 & 58.6 & 15.2 \\
    RAG \citep{petroni-etal-2021-kilt}            & 86.3 & 59.2  & 44.7 & 44.4 & 27.0 & 71.3 & 13.1 \\
    MT-DPR+BART \citep{maillard-etal-2021-multi}   & 86.3 & {-} & 58.0 & 39.8 & 31.8 & 59.6 & 15.3 \\
    MT-DPR+FiD \citep{piktus-etal-2021-web}    & 89.0 & 82.5  & 71.7 & 49.9 & 36.9 & 71.0 & 15.7 \\
    MT-DPR-WEB+FiD \citep{piktus-etal-2021-web} & 89.0 & 81.7  & 74.2 & 51.6 & 38.3 & \bfseries 72.7 & 15.5 \\ \midrule
    \system{}+FiD (LM+FM)          & 87.9 & 83.7  & 74.2 & 47.3 & 37.6 & 65.8 & 17.5 \\
    \system{}+FiD (LM+FM, intersective)             & \bfseries 89.5 & 83.6  & \bfseries 74.7 & \bfseries 53.7 & \bfseries 40.5 & 70.9 & 18.3 \\ \bottomrule
    \end{tabular}
    \caption{Downstream results on the KILT test set(s). Downstream metrics are accuracy (FEVER, T-REx, zero-shot RE), exact match (Natural Questions, HotpotQA, TriviaQA), or F1 (Wizard of Wikipedia). Best in bold. $\dag$: result taken from the \texttt{eval.ai} KILT leaderboard.}
    \label{tab:kilt-test-downstream}
\end{table*}
\begin{table}
\footnotesize
\centering
\begin{tabular}{ccS[table-format=2,round-precision=0]SS}
\toprule
{\textbf{System}} & 
{\textbf{Constr.}} & 
{\textbf{Beam}} & 
{\textbf{A@20}} &
{\textbf{A@100}}
\\
\midrule
SEAL    & \cmark  & 15   & 65.8 & 81.1 \\
(LM+FM)         & \xmark  & 15   & 65.3 & 80.1 \\
                 & \cmark  & 3    & 63.3 & 78.0 \\
                 & \cmark  & 5    & 64.7 & 79.9 \\
                 & \cmark  & 10   & 65.4 & 80.8 \\
\midrule
SEAL             & \cmark  & 15   & 76.2 & 86.3 \\
(LM+FM,          & \xmark  & 15   & 76.2 & 86.2 \\
intersective)      & \cmark  & 3    & 75.2 & 84.9 \\
                 & \cmark  & 5    & 75.9 & 85.8 \\
                 & \cmark  & 10   & 76.4 & 86.4 \\ 
\bottomrule
\end{tabular}
\caption{Ablation on Natural Questions. \system{} when using (\cmark) or not using (\xmark) FM-index constrained decoding, for beam size values in $\{3,5,10,15\}$. Reporting accuracy@$k$.}
\label{tab:ablation}
\end{table}
\begin{table*}[!ht]
\footnotesize
\centering
\setlength{\tabcolsep}{4pt}
\begin{tabular}{
    S[table-format=-3.1]
    r
    p{3.6cm}|
    p{4.7cm}
    p{4.7cm}
}
\toprule    
{\textbf{score}} & 
\# & 
{\textbf{identifier}} &
\multicolumn{1}{c}{\textbf{doc \#1}} & 
\multicolumn{1}{c}{\textbf{doc \#2}} 
\\ \midrule
273.2143157347409 & 1 &  earthquakes can be predicted 
&
\multirow{13}{*}{
\parbox{4.7cm}{
\textbf{Seismology @@} for precise \textbf{earthquake predictions}, including the VAN method. Most \textbf{seism}ologists do not believe that a system to provide timely warnings for individual \textbf{earthquakes} has yet been developed, and many believe that such a system would be unlikely to give \textbf{useful} warning of impending \textbf{seismic} events. However, more general \textbf{forecasts} routinely \textbf{predict} seismic \textbf{hazard}. Such \textbf{forecasts} \textbf{estimate} the \textbf{probability} of an \textbf{earthquake} of a particular [...]
}}
&
\multirow{13}{*}{
\parbox{4.7cm}{
\textbf{Earthquake prediction @@} reliably identified across significant spatial and temporal scales. While part of the scientific community hold that, taking into account non-\textbf{seismic} precursors and given enough resources to study them extensively, \textbf{prediction} might be \textbf{possible}, most scientists are pessimistic and some maintain that \textbf{earthquake prediction} is inherently impossible. \textbf{Predictions} are deemed significant if they can be shown to be successful beyond random chance.[...]
}}
\\
272.6594657747378 & 75 & Earthquake prediction @@ \\
269.9180385130301 & 3 &  predicted earthquakes \\
229.72512770800574 & 11 & Earthquake forecasting @@ \\
217.2180564605043 & 2 &  prediction Earthquake \\
211.45282257241695 & 1 &  used to predict earthquakes \\
205.31448885712462 & 7 &  earthquakes. Earthquake \\
 & {--} & &\\
-77.03246884558926 & 9 & Seismic metamaterial @@ \\
-97.40363285471005 & 14 & Seismic risk in Malta @@ \\
-113.43999116968067 & 3 & Quaternary (EP) @@ \\
-150.27171869350255 & 1 &  used to predict the locatio[...] \\
-301.4760316972612 & 17 & Precipice (Battlestar Gala[...] \\
\bottomrule 
\end{tabular}
\caption{Best (top) and worst (bottom) generated keys for the query ``can you predict earthquakes'' (left), and retrieved documents (right). Matched ngrams in bold. ``@@'' separates title and body.}
\label{tab:qualitative}
\end{table*}

\paragraph{Natural Questions}

We report in Table \ref{tab:vanilla-nq} the results of our evaluation on Natural Questions, a passage-level retrieval benchmark with a larger collection of documents (\ie, \textasciitilde$21$M w.r.t. $200$k in NQ320$k$). In this setting, the gap in performance between DSI-BART and \system{} is larger, possibly because memorizing documents identifiers in the parameters of the model becomes more challenging with larger corpora. 
Remarkably, the intersective formulation of \system{} achieves results comparable or superior to more established retrieval paradigms (\eg, BM25, DPR and GAR). 
To better understand the generalization capabilities of our retrieval solution we use the question/answer overlap split of~\citet{lewis-etal-2021-question}. This study reveals that \system{} achieves the highest performance for question/answer pairs never seen during training (\ie, no overlap), suggesting a better ability to generalize to completely
novel questions with novel answers (\eg, $3.5$ points better than GAR on average). 

\paragraph{KILT}
We report retrieval results at passage level on the KILT benchmark in Table \ref{tab:kilt-dev-rp}.\footnote{We report page-level and KILT-score results in the Appendix (§\ref{sec:additional-kilt}).} \system{} outperforms DPR by more than 10  points on average in passage-level R-precision, indicating that our method is more precise in surfacing ground truth evidence as the first result. 
Moreover, \system{} also performs better than MT-DPR (multi-task DPR) even when the latter is pretrained on tens of millions of questions from PAQ~\citep{lewis-etal-2021-paq}, a technique that can drastically improve results and that could potentially bring benefits to our method as well (a task we leave for future work).
When it comes to downstream performances  (Table \ref{tab:kilt-test-downstream}), FiD with passages retrieved by intersective \system{} establishes a new state-of-the-art on 4 datasets out of 7 (FEVER, zsRE, NQ, HoPo), and achieves very competitive results on the remaining 3. 

\paragraph{Speed and constrained decoding} The inference speed of \system{} is directly proportional to the beam size, with a limited overhead added by constrained decoding. On the Natural Questions test set, for instance, retrieval with the intersective scoring requires on our 1 GPU evaluation setup \textasciitilde16~minutes and \textasciitilde35~minutes with, respectively, a beam size of $5$ or $15$.~\citet{mao-etal-2021-generation} report a lower runtime for GAR (\textasciitilde5~minutes), and a comparable one for DPR (\textasciitilde30~minutes). 
Note that more efficient approaches to constrained decoding have been proposed (\eg,~\citet{de-cao-etal-2021-highly}) and we leave their application to \system{} as future work. 

\paragraph{Ablation studies}
In Table \ref{tab:ablation} we report performances on Natural Questions for various configurations of \system{}.
While, in general, performances increase with a larger beam, diminishing returns (or even a slight performance decrease) are encountered between a value of $10$ and $15$. Disabling constrained decoding and discarding a posteriori all generated ngrams that don't appear in the corpus,
results in slightly lower performances. 

\paragraph{Qualitative Analysis}
In Table \ref{tab:qualitative}, we show examples of ngrams predicted by \system{} (trained on KILT) given the query ``can you predict earthquakes''. \system{} is able to rephrase
the query in ways that preserve its lexical material producing ngrams such as \textit{earthquakes can be predicted}, \textit{used to predict earthquakes} etc. Morevoer, the model is also able to explore more diverse regions of the output space, overcoming the vocabulary mismatch problem: ngrams contain related tokens like the subword \textit{seism-} and the word \textit{forecast}. \system{}'s LM+FM scoring is also able to assign a score below $0$ (and, thus, exclude from the search), unrelated ngrams that are considered by the beam because of their promising start, such as ``Seismic risk in Malta @@''. 

\section{Discussion}
With \system{} we present solution that could potentially find applications outside information retrieval (\eg, enforce generated substrings come from a white list of trusted sources).
While we conduct our experiments with a model of \textasciitilde$400$M parameters (\ie, BART) for fast iterations, we believe the use of larger models could considerably improve performance. Changing the model would not affect the size of the index nor the cost of using it --- $O(|n|\text{log}|V|)$ for finding an ngram $n$. 
Moreover, we believe that indexing very large corpora (\eg, the web) could be done more efficiently than existing attempts (\eg, \citet{piktus-etal-2021-web}) given the light memory footprint. Finally, dynamic variants~\citep{Gerlach2007DynamicFF,Salson2009a} could allow the update of the FM-index on the fly without the need of re-indexing. While out of the scope of the current paper, we plan to tackle some of these scaling challenges in future work.

\section{Conclusion}
In this paper we present \system{}, a novel retrieval system that combines an autoregressive language model with a compressed full-text substring index. Such combination allows to constraint the generation of existing ngrams in a corpus and to jointly retrieve all the documents containing them.
Empirically, we show an improvement of more than $10$ points in average passage-level R-precision on KILT, and establish new state-of-the-art downstream performance on 4 out 7 datasets when paired with a reader model. 
While our results show that \system{} could already compete with more established retrieval systems, we believe there is potential in exploring the use of existing (or yet to come) larger autoregressive models.

\section*{Acknowledgements}
We thank Aleksandra Piktus, Edoardo Barba, Niccolò Campolungo, and Pere-Lluis Huguet Cabot for their helpful comments and suggestions.

\bibliography{anthology,custom}
\bibliographystyle{acl_natbib}

\FloatBarrier

\appendix

\begin{table*}[t]
\footnotesize
\setlength{\tabcolsep}{5pt}
\centering
\begin{tabular}{lSSSSSSS|S}
\toprule
\textbf{Model}        & {\textbf{FEV}}  & {\textbf{T-REx}} & {\textbf{zsRE}} & {\textbf{NQ}}   & {\textbf{HoPo}} & {\textbf{TQA}}  & {\textbf{WoW}}  & {\textbf{AVG}}  \\ \midrule
KGI \citep{glass-etal-2021-robust}               
& 75.60 & 74.36 & \bfseries 98.49 & 63.71 & {-}   & 60.49 & 55.37 & {-} \\
Hindsight \citep{paranjape-etal-2021-hindsight}         
& {-}   & {-}   & {-}   & {-}   & {-}   & {-}   & 56.08 & {-} \\
GENRE \citep{decao-etal-2021-autoregressive}             
& \bfseries 83.64 & \bfseries 79.42 & 95.81 & 60.25 & 51.27 & \bfseries 69.16 & \bfseries 62.88 & \bfseries 71.8 \\
MT-DPR \citep{maillard-etal-2021-multi}
& 74.5 & 69.5 & 80.9 & 59.4 & 42.9 & 61.5 & 41.1 & 61.4 \\
MT-DPR+WEB \citep{piktus-etal-2021-web}        
& 74.77 & 75.64 & 89.65 & 59.83 & 45.38 & 58.85 & 41.54 & 63.7  \\ \midrule
\system{} (LM+FM)
& 77.8 & 67.8 & 98.0 & 60.3 & 54.0 & 68.1 & 55.4 & 68.8 \\
\system{} (LM+FM, intersective)	
& 81.4 & 62.1 & 91.6 & \bfseries 63.2 & \bfseries 58.8 & 68.4 & 57.5 & 69.0 \\
\bottomrule
\end{tabular}
\caption{Retrieval results on the KILT test set(s). Reporting page-level R-precision (higher is better). Best in bold. Results are taken from the \texttt{eval.ai} KILT leaderboard.}
\label{tab:kilt-test-rp-page}
\end{table*}
\begin{table*}[t]
\footnotesize
    \centering
    \begin{tabular}{lSSSSSSS}
    \toprule
    \multirow{2}{*}{\textbf{System}}         & {\textbf{FEV}}  & {\textbf{T-REx}} & {\textbf{zsRE}} & {\textbf{NQ}}   & {\textbf{HoPo}} & {\textbf{TQA}}  & {\textbf{WoW}}  \\ 
                   & {\textsc{k.-acc}}  & {\textsc{k.-acc}}   & {\textsc{k.-acc}}  & {\textsc{k.-em}}   & {\textsc{k.-em}}   & {\textsc{k.-em}}   & {\textsc{k.-f1}}   \\ \midrule

KGI \citep{glass-etal-2021-robust} & 64.4 & \bfseries 69.1 & 72.3 & 36.4 & {-} & 42.9 & 10.4 \\
Hindsight \citep{paranjape-etal-2021-hindsight} & {-} & {-} & {-} & {-} & {-} & {-} & \bfseries 13.4 \\
RAG \citep{petroni-etal-2021-kilt} & 53.5 & 23.1 & 36.8 & 32.7 & 3.2 & 38.1 & 8.8 \\
MT-DPR+BART \citep{maillard-etal-2021-multi} & 63.9 & {-} & 50.6 & 29.1 & 9.5 & 42.4 & 5.9 \\
MT-DPR-WEB+FiD \citep{piktus-etal-2021-web} & 65.7 & 64.6 & 67.2 & 35.3 & 11.7 & 45.6 & 7.6 \\
\midrule
\system{}+FiD (LM+FM) & 67.0 & 60.1 & \bfseries 73.2 & 32.8 & 15.1 & 47.7 & 11.0 \\
\system{}+FiD (LM+FM, intersective) & \bfseries 71.3 & 54.6 & 69.2 & \bfseries 38.8 & \bfseries 18.1 & \bfseries 50.6 & 11.6 \\
\bottomrule
\end{tabular}
\caption{KILT scores on the KILT test set(s). In KILT-scores an instance is considered correct if both the predicted page and the answer match the ground truth. Metrics are accuracy (FEVER, T-REx, zero-shot RE), exact match (Natural Questions, HotpotQA, TriviaQA), or F1 (Wizard of Wikipedia). Best in bold. Results are taken from the \texttt{eval.ai} KILT leaderboard.}
    \label{tab:kilt-test-kilt-scores}
\end{table*}
\FloatBarrier

\section{Additional KILT results}
\label{sec:additional-kilt}
We report in Table \ref{tab:kilt-test-rp-page} page-level results on the KILT test set. On most datasets, \system{} obtains results which are comparable or better than other systems performing page-level retrieval. Furthermore, are results are within two points of the average performance of GENRE, \ie, a system that directly targets the page-level setting. Comparing KILT-scores (Table \ref{tab:kilt-test-kilt-scores}), \ie, a metric combining downstream performances and page-level R-precision, we achieve state-of-the-art results on 4 out of 7 datasets.

\section{Impact of unsupervised examples}
\system{} is trained with both supervised and unsupervised examples. In Table \ref{tab:nq-ablation-data} we report ablated results, by which we assess the importance of both kind of training examples. The addition of unsupervised examples improves purely supervised training by one point (A@100). Only training with unsupervised examples results in performances which are slightly below BM25's.
\begin{table}[h]
    \centering
    \begin{tabular}{cccSS}
        \toprule
        \textbf{System} & \textbf{Sup.} & \textbf{Unsup.} & {\bfseries A@20} & {\bfseries A@100}  \\
        \midrule
        BM25 & - & - & 62.9 & 78.1 \\
        \midrule
        \system{} & \cmark & \cmark & 76.2 & 86.3\\
        (LM+FM & \cmark & \xmark & 74.81 & 85.4 \\
        intersective) & \xmark & \cmark & 61.71 & 76.32 \\
        \bottomrule
    \end{tabular}
    \caption{Ablation on Natural Questions. \system{} when using (\cmark) or not using (\xmark) supervised/unsupervised data. Reporting accuracy@$k$.}
    \label{tab:nq-ablation-data}
\end{table}

\end{document}